\documentclass[runningheads]{llncs}
\usepackage{graphicx}
\usepackage{array}
\usepackage{multirow}
\usepackage{amsmath,amssymb} 
\usepackage{color}
\usepackage[nosort, nocompress]{cite}
\usepackage[width=122mm,left=12mm,paperwidth=146mm,height=193mm,top=12mm,paperheight=217mm]{geometry}
\DeclareMathOperator{\E}{\mathbb{E}}
\DeclareMathOperator*{\argmin}{argmin} 
\newcommand{\overbar}[1]{\mkern 5mu\overline{\mkern-5mu#1\mkern-5mu}\mkern 5mu}
\newcommand{\calcfactor}[1]{%
  \dimexpr#1\textwidth-2\tabcolsep-1.5\arrayrulewidth\relax
}
\newcolumntype{P}[1]{p{\calcfactor{#1}}}

\newcommand{\keypoint}[1]{\noindent\textbf{#1}\quad}

\begin{document}
\title{Deep Factorised Inverse-Sketching} 

\titlerunning{Deep Factorised Inverse-Sketching}
%
\author{Kaiyue Pang\inst{1} \and
Da Li\inst{1} \and
Jifei Song\inst{1} \and
Yi-Zhe Song\inst{1} \and
Tao Xiang\inst{1} \and
Timothy M. Hospedales\inst{1,2}
}
%
\authorrunning{Kaiyue Pang et al.}
%

\institute{SketchX, Queen Mary University of London, London, UK \\
\email{\{kaiyue.pang, da.li, j.song, yizhe.song, t.xiang\}@qmul.ac.uk}\\ \and
The University of Edinburgh, Edinburgh, UK\\
\email{t.hospedales@ed.ac.uk}}
\maketitle              
\begin{abstract}
Modelling human free-hand sketches has become topical recently, driven by practical applications such as fine-grained sketch based image retrieval (FG-SBIR). Sketches are clearly related to photo edge-maps, but a human free-hand sketch of a photo is not simply a clean rendering of that photo's edge map. Instead there is a fundamental process of abstraction and iconic rendering, where overall geometry is warped and salient details are selectively included. In this paper we study this sketching process and attempt to invert it. We model this inversion by translating iconic free-hand sketches to contours that resemble more geometrically realistic projections of object boundaries, and separately factorise out the salient added details. This factorised re-representation makes it easier to match a free-hand sketch to a photo instance of an object. Specifically, we propose a novel unsupervised image style transfer model based on enforcing a cyclic embedding consistency constraint. A deep FG-SBIR model is then formulated to accommodate complementary discriminative detail from each factorised sketch for better matching with the corresponding photo. Our method is evaluated both qualitatively and quantitatively to demonstrate its superiority over a number of state-of-the-art alternatives for style transfer and FG-SBIR.
%
\end{abstract}
\section{Introduction}

Free-hand sketch is the simplest form of human visual rendering. Albeit with varying degrees of skill, it comes naturally to humans at young ages, and has been used for millennia. Today it provides a convenient tool for communication, and a promising input modality for visual retrieval. Prior sketch studies focus on sketch recognition \cite{eitz2012humans} or sketch-based image retrieval (SBIR). SBIR methods can be further grouped into category-level \cite{eitz2011sketch} and instance-level fine-grained SBIR (FG-SBIR) \cite{yu2016sketch}. This dichotomy corresponds to how a sketch is created --  based on a category-name or a (real or mental) picture of a specific object instance. These produce different granularities of visual cues (e.g., prototypical vs. specific object detail). As argued in \cite{yu2016sketch}, it is fine-grained sketches of specific object instances that bring practical benefit for image retrieval over the standard text modality.

\begin{figure}[t]
\centering
\includegraphics[width=0.85\textwidth]{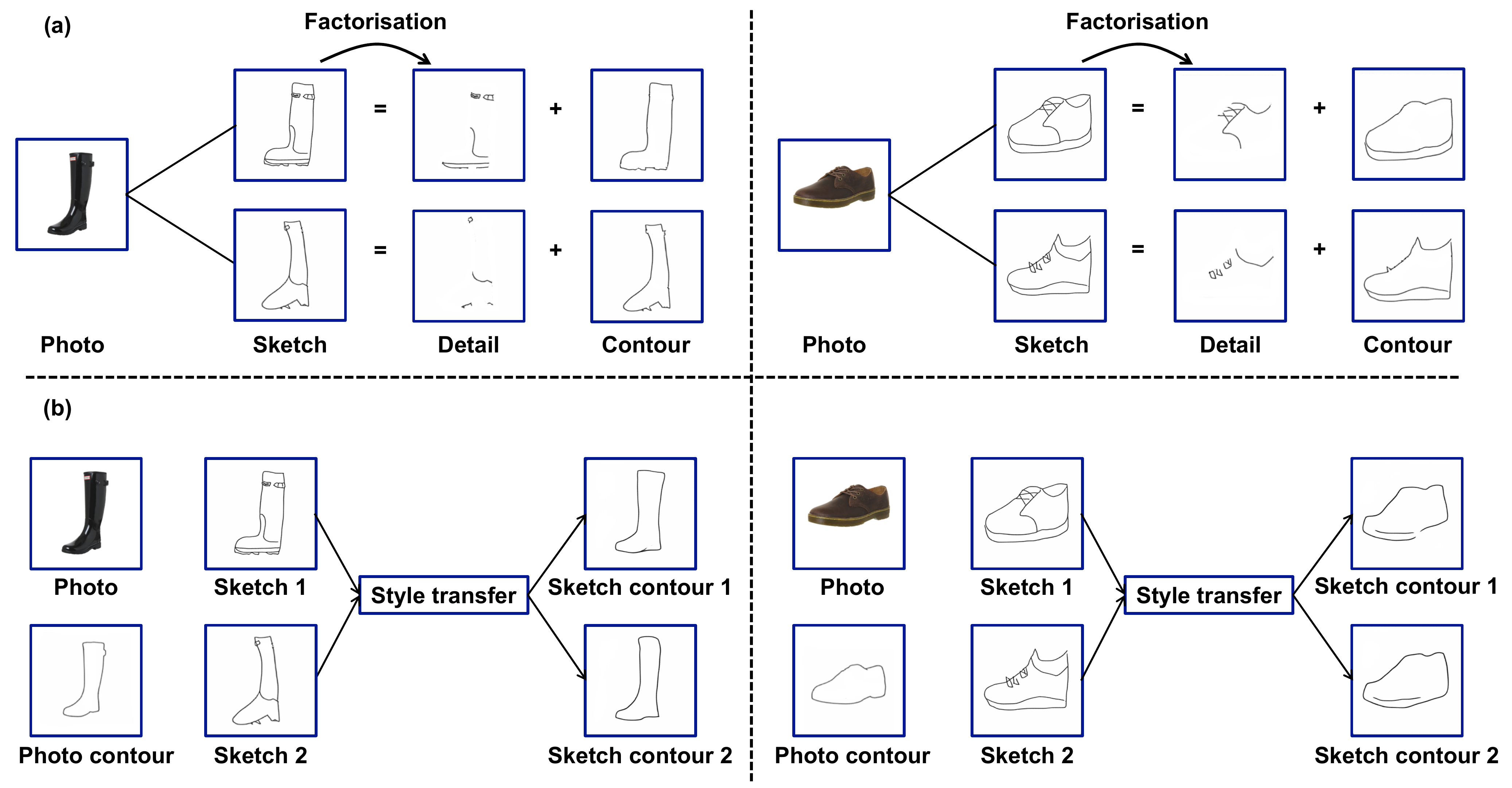}
\caption{(a) A free-hand object instance sketch consists of two parts: iconic contour and object details. (b) Given a sketch, our style transfer model restyles it into distortion-free contour. The synthesised contours of different sketches of the same object instance resembles each other as well as the corresponding photo contour.}
\label{fig:thumbnail}
\end{figure}

Modelling fine-grained object sketches and matching them with corresponding photo images containing the same object instances is extremely challenging. This is because photos are exact perspective projections of a real world scene or object, while free-hand sketches are iconic abstractions with different geometry, and selected choice of included detail. Moreover, sketches are drawn by people of different backgrounds, drawing abilities and styles, and different subjective perspectives about the salience of details to include. Thus two people can draw very different sketches of the same object as shown in Fig.~\ref{fig:thumbnail}(a) photo$\to$sketch.

A closer inspection of the human sketching process reveals that it includes two components. As shown in \cite{li2017free}, a sketcher typically first deploys long strokes to draw iconic object contours, followed by shorter strokes to depict visual details (e.g., shoes laces or buckles in Fig.~\ref{fig:thumbnail}(a)). Both the iconic contour and object details are important for recognising the object instance and matching a sketch with its corresponding photo. The contour is informative about object subcategory (e.g., a boot or trainer), while the details distinguish instances within the subcategory -- modelling both are thus necessary. However, they have very different characteristics demanding different treatments. The overall geometry of the sketch contour  experiences large and user-specific distortion compared to the true edge contour of the photo (compare sketch contour in Fig.~\ref{fig:thumbnail}(a) with photo object contour in  Fig.~\ref{fig:thumbnail}(b)). Photo edge contours are an exact perspective projection of the object boundary; and free-hand sketches are typically an orthogonal projection at best, and usually much more distorted than that -- if only because humans seem unable to draw long smooth lines without distortion \cite{FlashHogan1985}. In contrast, distortion is less of an issue for shorter strokes in the object detail part. But choice and amount of details varies by artist (e.g., buckles in Fig.~\ref{fig:thumbnail}(a)).

In this paper, for the first time, we propose to model human sketches by inverting the sketching process. That is, instead of modelling the forward sketching pass (i.e., from photo/recollection to sketch), we study the inverse problem of translating sketches into visual representations that closely resemble the perspective geometry of photos. We further argue that this inversion problem is best tackled on two levels by separately factorising out object contours and the salient sketching details. Such factorisation is important for both modelling sketches and matching them with photos. This is due to the differences mentioned above: sketch contours are consistently present but suffer from large distortions, while details are less distorted but more inconsistent in their presence and abstraction level. Both parts can thus only be modelled effectively when they are factorised. 

We tackle the first level of inverse-sketching by proposing a novel deep image synthesis model for style transfer. It takes a sketch as input, restyles the sketch into natural contours resembling the more geometrically realistic contours extracted from photo images, while removing object details (see Fig.~\ref{fig:thumbnail}(b)). This stylisation task is extremely difficult because (a) Collecting a large quantity of sketch-photo pairs is infeasible so the model needs to be trained in an unsupervised manner. (b) There is no pixel-to-pixel correspondence between the distorted sketch contour and realistic photo contour, making models that rely on direct pixel correspondence such as \cite{isola2016image} unsuitable. To overcome these problems, we introduce a new cyclic embedding consistency in the proposed unsupervised image synthesis model. It forces the sketch and unpaired photo contours to share some support in a common low-dimensional semantic embedding space.

We next complete the inversion in a discriminative model designed for matching sketches with photos. It importantly utilises the synthesised contours to factor out object details to better assist with sketch-photo matching. Specifically, given a training set of sketches, their synthesised geometrically-realistic contours, and corresponding photo images, we develop a new FG-SBIR model that extracts factorised feature representations corresponding to the contour and detail parts respectively before fusing them to match against the photo. The model is a deep Siamese neural network with four branches. The sketch and its synthesised contours have their own branches respectively. A decorrelation loss is applied to ensure the two branch's representations are complementary and non-overlapping (i.e., factorised). The two features are then fused and subject to triplet matching loss with the features extracted from the positive and negative photo branches to make them discriminative.

The contributions of this work are as follows: (1) For the first time, the problem of factorised inverse-sketching is defined and identified as a key for both sketch modelling and sketch-photo matching. (2) A novel unsupervised sketch style transfer model is proposed to translate a human sketch into a geometrically-realistic contour. (3) We further develop a new FG-SBIR model which extracts an object detail representation to complement the synthesised contour for effective matching against photos.

\section{Related Work}

\keypoint{Sketch modelling:} There are several lines of research aiming to deal with abstract sketches so that either sketch recognition or SBIR can be performed. The most best studied is invariant representation engineering or learning. These either aim to hand-engineer features that are invariant to abstract sketch vs concrete photo domain \cite{eitz2011sketch, hu2013evalSKBIR, bui2015scalable}, or learn a domain invariant representation given supervision of sketch-photo categories \cite{liu2017deep,shen2018zero,hu2018sketch} and sketch-photo pairs \cite{yu2016sketch,sangkloy2016sketchy}. More recent works have attempted to leverage insights from the human sketching process. \cite{berger2013style,yu2017sketch} recognised the importance of stroke ordering, and \cite{yu2017sketch} introduced ordered stroke deformation as a data augmentation strategy to generate more training sketches for sketch recognition task. The most explicit model of sketching to our knowledge is the stroke removal work considered in \cite{umar2018sketch}. It abstracts sketches by proposing reinforcement learning (RL) of a stroke removal policy that  estimates which strokes can be safely removed without affecting recognisability. It evaluates on FG-SBIR and uses the proposed RL-based framework to generate abstract variants of training sketches for data augmentation. Compared to \cite{yu2017sketch} and \cite{umar2018sketch}, both of which perform within-domain abstraction (i.e., sketch to abstracted sketch), our approach presents a fundamental shift in that it models the inverse-sketching process (i.e., sketch to photo contour) therefore directly solving for the sketch-photo domain gap, without the need for data augmentation. Finally, we note that no prior work has taken our step of modelling sketches by factorisation into contour and detail parts.

\keypoint{Neural image synthesis:} Recent advances in neural image synthesis have led to a number of practical applications, including image stylisation \cite{gatys2016image, johnson2016perceptual,luan2017deep, liao2017visual}, single image super-resolution \cite{ledig2017photo}, video frame prediction \cite{mathieu2015deep}, image manipulation \cite{zhu2016generative, korshunova2017fast} and conditional image generation \cite{mirza2014conditional, yan2016attribute2image, reed2016generative, odena2017conditional, zhang2017stackgan}. 
The models most relevant to our style transfer model are deep image-to-image translation models \cite{isola2016image,zhu2017unpaired, liu2017unsupervised, kim2017learning, yi2017dualgan}, particularly the unsupervised ones \cite{zhu2017unpaired, liu2017unsupervised, kim2017learning, yi2017dualgan}. The goal is to translate an image from one domain to another with a deep encoder-decoder architecture. In order to deal with the large domain gap between a sketch containing both distorted sketch contour and details and a distortion-free contour extracted from photo edges, our model has a novel component, that is, instead of the cyclic visual consistency deployed in \cite{zhu2017unpaired, liu2017unsupervised, kim2017learning, yi2017dualgan}, we enforce cyclic embedding constraint, a softer version for better synthesis quality. Both qualitative and quantitative results show that our model outperforms existing models.

\keypoint{Fine-grained SBIR:} In the context of image retrieval, sketches provide a convenient modality for providing fine-grained visual query descriptions --- a sketch speaks for a `hundred' words. FG-SBIR was first proposed in \cite{li2014fine}, which employed a deformable part-based model (DPM) representation and graph matching. It is further tackled by deep models \cite{yu2016sketch,sangkloy2016sketchy,song2017sketch} which aim to learn an embedding space where sketch and photo can be compared directly -- typically using a three-branch  Siamese network with a triplet ranking loss. More recently, FG-SBIR was addressed from an image synthesis perspective \cite{pang2017fgsbir} as well as an explicit photo to vector sketch synthesis perspective \cite{song2018sketch}. The latter study used a CNN-RNN generative sketcher and used the resulting synthetic sketches for data augmentation. 
Our FG-SBIR model is also a Siamese joint embedding model. However, it differs in that it employs our synthesised distortion-free contours both as a bridge to narrow the domain gap between sketch and photo, and a means for factorising out the detail parts of the sketch. We show that our model is superior to all existing models on the largest FG-SBIR dataset.

\section{Sketch Stylisation with Cyclic Embedding Consistency}
\keypoint{Problem definition:}
Suppose we have a set of free-hand sketches $S$ drawn by amateurs based on their mental recollection of object instances \cite{yu2016sketch} and a set of photo object contours $C$ sparsely extracted from photos using an off-the-shelf edge detection model\cite{ZitnickECCV14edgeBoxes}, with empirical distribution $s \sim p_{data}(S)$ and $c \sim p_{data}(C)$ respectively. They are theme aligned but otherwise \emph{unpaired} and  \emph{non-overlapped} meaning they can contain different sets of object instances. This makes training data collection much easier. Our objective is to learn an unsupervised deep style transfer model, which inverts the style of a sketch to a cleanly rendered object contour with more realistic geometry, and user-specific details removed (see Fig.~\ref{fig:thumbnail}(b)).

\begin{figure}[t]
\centering
\includegraphics[width=0.85\textwidth]{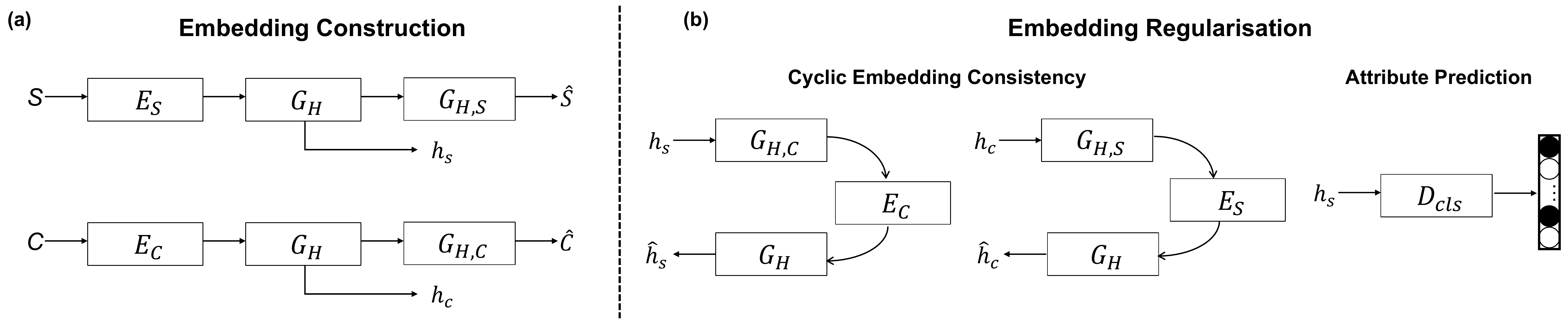}
\caption{Schematic of our sketch style transfer model with cyclic embedding consistency. (a) Embedding space construction. (b) Embedding regularisation through cyclic embedding consistency and an attribute prediction task. }
\label{fig:embedding}
\end{figure}

\subsection{Model Formulation}\label{sec:stylisation model}

Our model aims to transfer images in a source domain (original human sketches) to a target domain (photo contours). It consists of two  encoder-decoders, $\{E_S, G_S\}$ and $\{E_C, G_C\}$, which map an image from the source (target) domain to the target (source) domain and produce an image whose style is indistinguishable from that in the target (source) domain. Once learned, we can use $\{E_S, G_C\}$ to transfer the style of $S$ into that of $C$, i.e., distortion-free and geometrically realistic  contours.  Note that under the unsupervised (unpaired) setting, such a mapping is highly under-constrained 
-- there are infinitely many mappings $\{E_S, G_C\}$ that will induce the same distribution over contours $c$. This issue calls for adding more structural constraints into the loop, to ensure $s$ and $c$ lie on some shared embedding space for effective style transfer and instance identity preserving between the two. To this end, the decoder $G_S$ ($G_C$) is decomposed into two sub-networks: a shared embedding space construction subnet $G_H$, and an unshared embedding decoder $G_{H,S}$ ($G_{H,C}$), i.e., $G_S \equiv G_H \circ G_{H,S}, G_C \equiv G_H \circ G_{H,C}$ (see Fig.~\ref{fig:embedding}(a)).   

\keypoint{Embedding space construction:} We construct our embedding space similarly to  \cite{liu2016coupled, liu2017unsupervised}: The $G_H$ projects the outputs of the encoders into a shared embedding space. We thus have $h_s = G_H(E_S(s)), h_c = G_H(E_C(c))$. The projections in the embedding space are then used as inputs by the  decoder to perform reconstruction:   $\hat{s} = G_{H,S}(h_s), \hat{c} = G_{H,C}(h_c)$.

\keypoint{Embedding regularisation:} As illustrated in Fig.~\ref{fig:embedding} (b), the embedding space is learned with two regularisations: (i) Cyclic embedding consistency: this exploits the property that the learned style transfer should be `embedding consistent', that is, given a translated image, we can arrive at the same spot in the shared embedding space with its original input. This regularisation is formulated as $h_{s}=G_H(E_S(s))\rightarrow G_{H,C}(G_H(E_S(s)))\rightarrow G_H(E_C(G_{H,C}(G_H(E_S(s)))))\approx h_{s}$, and $h_{c}=G_H(E_C(c))\rightarrow G_{H,S}(G_H(E_C(c)))\rightarrow G_H(E_S(G_{H,S}(G_H(E_C(c))))) \approx h_{c}$ for the two domains respectively. This is  different from the  cyclic visual consistency used by existing unsupervised image-to-image translation models\cite{liu2016coupled, liu2017unsupervised,zhu2017unpaired}, by which the input image is reconstructed by translating back the translated input image. The proposed cyclic embedding consistency is much `softer' compared to the cyclic visual consistency since the reconstruction is performed in the embedding space rather than at the per-pixel level in the image space. It is thus more capable of coping with domain discrepancies caused by the large pixel-level mis-alignments due to contour distortion and the missing of details inside the contours.  (ii) Attribute prediction: to cope with the large variations of sketch appearance when the same object instance is drawn by different sketchers (see Fig.~\ref{fig:thumbnail}(a)), we add an attribute prediction task to the embedding subnet so that the embedding space needs to preserve all the information required to predict a set of semantic attributes.

\begin{figure}[t]
\centering
\includegraphics[width=0.85\textwidth]{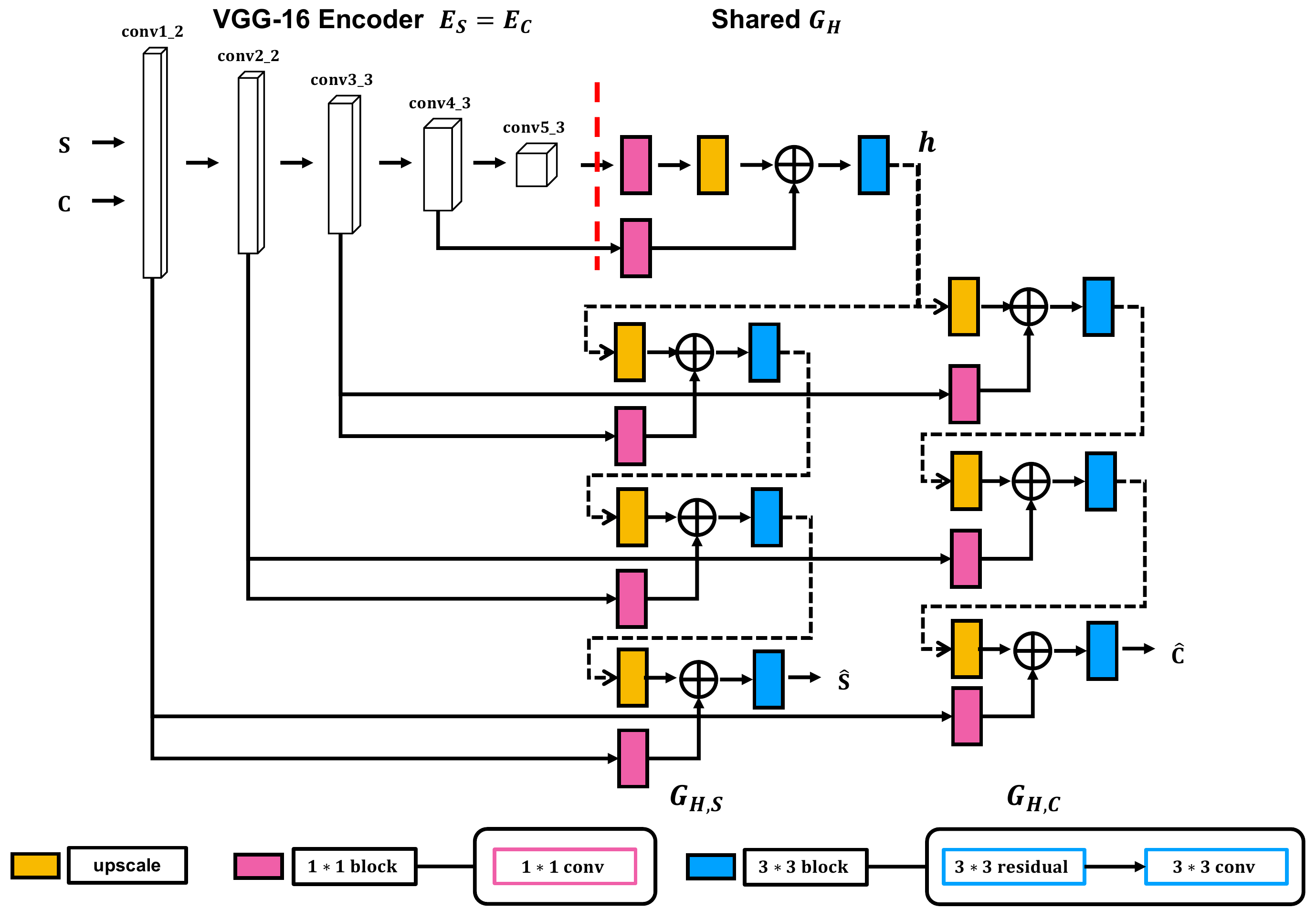}
\caption{A schematic of our specifically-designed encoder-decoder. }
\label{fig:framework}
\end{figure}

\keypoint{Adversarial training:} Finally, as in most existing deep image synthesis models, we introduce a discriminative network to perform adversarial training \cite{goodfellow2014generative}: the discriminator is trained to be unable to distinguish generated contours from sketch inputs and the photo contours extracted from object photos. 

\subsection{Model Architecture}
\keypoint{Encoder:} Most existing unsupervised image-to-image translation models  design a specific encoder architecture and train the encoder from scratch. We found that this works poorly for sketches due to  lack of training data and the large appearance variations mentioned earlier. We therefore adopt a fixed VGG encoder pretrained on ImageNet. As shown in Fig.~\ref{fig:framework}, the encoder consists of five convolutional layers before each of the five max-pooling operations of a pre-trained VGG-16 network, namely $conv1\_ 2$, $conv2\_ 2$, $conv3\_ 3$, $conv4\_ 3$ and $conv5\_ 3$.  Note that adopting a pretrained encoder means that now we have $E_S = E_C$.

\keypoint{Decoder:} The two subnets of the decoder: $G_H$ and $G_{H,S}$ ($G_{H,C}$) use a residual design. Specifically, for convolutional feature map extracted at each spatial resolution, we start with $1 * 1$ conv, upsample it by a factor of 2 with bilinear interpolation and then add the output of the corresponding encoder layer.  It is further followed by a $3 * 3$ residual and $3 * 3$ conv for transformation learning and adjusting appropriate channel numbers for the next resolution. Note that shortcut connections between the encoder and decoder corresponding layers are also established in the residual form.  As illustrated in Fig.~\ref{fig:framework}, the shared embedding construction subnet $G_H$ is composed of one such block while the unshared embedding decoders $G_{H,S}$ ($G_{H,C}$) have three. For more details of the encoder/decoder and discriminator architecture, please see Sec.~\ref{sec:exp setting}.

\subsection{Learning Objectives}
\keypoint{Embedding consistency loss:} Given  $s$ ($c$), and its cross-domain synthesised image $G_C(E_S(s))$ ($G_S(E_C(c))$), they should arrive back to the same location in the embedding space. We enforce this by  minimising the Euclidean distance between them in the embedding space:
\begin{equation}
\label{equ:embedding_loss}
\begin{aligned}
\mathcal{L}_{embed} = \E_{s\sim S, c\sim C}[&||G_H(E_S(s))-G_H(E_C(G_C(E_S(s))))||_{2} \\ + &||G_H(E_C(c))-G_H(E_S(G_S(E_C(c))))||_{2}].
\end{aligned}
\end{equation}

\keypoint{Self-reconstruction loss:} Given  $s$ ($c$), and its  reconstructed result $G_S(E_S(s))$ ($G_C(E_C(c))$), they should be visually close. We thus have
\begin{equation}
\label{equ:recons_loss}
\begin{aligned}
\mathcal{L}_{recons} = \E_{s\sim S, c\sim C}[||s-G_S(E_S(s))||_{1} + ||c-G_C(E_C(c))||_{1}].
\end{aligned}
\end{equation}

\keypoint{Self-reconstruction loss:} Given  $s$ ($c$), and its  reconstructed result $G_S(E_S(s))$ ($G_C(E_C(c))$), they should be visually close. We thus have
\begin{equation}
\label{equ:recons_loss}
\begin{aligned}
\mathcal{L}_{recons} = \E_{s\sim S, c\sim C}[||s-G_S(E_S(s))||_{1} + ||c-G_C(E_C(c))||_{1}].
\end{aligned}
\end{equation}

\keypoint{Attribute prediction loss:} Given a sketch $s$ and its semantic attribute vector $a$, we hope its embedding  $G_H(E_S(s))$ can be used to predict the attributes $a$. To realise this, we introduce an auxiliary one-layer subnet $D_{cls}$ on top of the embedding space $h$ and minimise the classification errors:
\begin{equation}
\label{equ:cls_loss}
\begin{aligned}
\mathcal{L}_{cls} = \E_{s,a \sim S}[-\log D_{cls}(a|G_H(E_S(s)))].
\end{aligned}
\end{equation}
\keypoint{Domain-adversarial loss:} Given $s$ ($c$) and its cross-domain synthesised image $G_C(E_S(s))$ ($G_S(E_C(c))$), the synthesised image should be indistinguishable to a target domain image $c$ ($s$) using the adversarially-learned discriminator, denoted $D_C$ ($D_S$). To stabilise training and improve the quality of the synthesised images, we adopt the least square generative adversarial network (LSGAN) \cite{mao2017least} with gradient penalty \cite{gulrajani2017improved}. The  domain-adversarial loss is defined as:
\begin{align}
\label{equ:advg_loss}
\begin{aligned}
\mathcal{L}_{adv_{g}} & = \E_{s \sim S}[||D_C(G_C(E_S(s)))-1||_2] \\& + \E_{c \sim C}[||D_S(G_S(E_C(c)))-1||_2] \\
\mathcal{L}_{adv_{ds}} &= \E_{s \sim S}[||D_S(s)-1||_2]+ \E_{c \sim C}[||D_S(G_S(E_C(c)))||_2] \\& -\lambda_{gp}\E_{\tilde{s}}[(||\nabla_{\tilde{s}}D_S(\tilde{s})||_2 - 1)^2]\\
\mathcal{L}_{adv_{dc}} &= \E_{c \sim C}[||D_C(c)-1||_2]+ \E_{s \sim S}[||D_C(G_C(E_S(s)))||_2] \\& -\lambda_{gp}\E_{\tilde{c}}[(||\nabla_{\tilde{c}}D_C(\tilde{c})||_2 - 1)^2] 
\end{aligned}
\end{align}
\noindent where $\tilde{s}, \tilde{c}$ are sampled uniformly along a straight line between their corresponding domain pair of real and generated images. We set weighting factor $\lambda_{gp}=10$.

\keypoint{Full learning objectives:} Our full model is trained alternatively as with a standard conditional GAN framework, with the following joint optimisation:
\begin{equation}
\label{equ:full_loss}
\begin{aligned}
\argmin_{D_S,D_C}\lambda_{adv}L_{adv_{ds}} &+ \lambda_{adv}L_{adv_{dc}}\\
\argmin_{E_{S}, E_{C}, G_{S}, G_{C}, D_{cls}} \lambda_{embed}L_{embed} &+ \lambda_{recons}L_{recons} +\lambda_{adv}L_{adv_g} + \lambda_{cls}L_{cls}
\end{aligned}
\end{equation}
\noindent where $\lambda_{adv},\lambda_{embed},\lambda_{recons}, \lambda_{cls}$ are hyperparameters that control the relative importance of each loss. In this work, we set $\lambda_{adv}=10,\lambda_{embed}=100,\lambda_{recons}=100$ and $\lambda_{cls}=1$ to  keep the losses in roughly the same value range.

\section{Discriminative Factorisation for FG-SBIR}
\label{sec:fgsbir}

The sketch style transfer model in Sec.~\ref{sec:stylisation model} addresses the first level of inverse-sketching by translating a sketch into a geometrically realistic contour. Specifically, for a given sketch $s$, we can synthesise its distortion-free sketch contour $s_c$ as $G_C(E_S(s))$. However, the model is not trained to synthesise the sketch details inside the contour -- this is harder because sketch details exhibit more subjective abstraction yet less distorted.
In this section, we show that for learning a discriminative FG-SBIR model, such a partial factorisation is enough: we can take $s$ and $s_c$ and extract complementary detail features from $s_c$ to complete the inversion process.

\begin{figure}[t]
\centering
\includegraphics[width=0.85\textwidth]{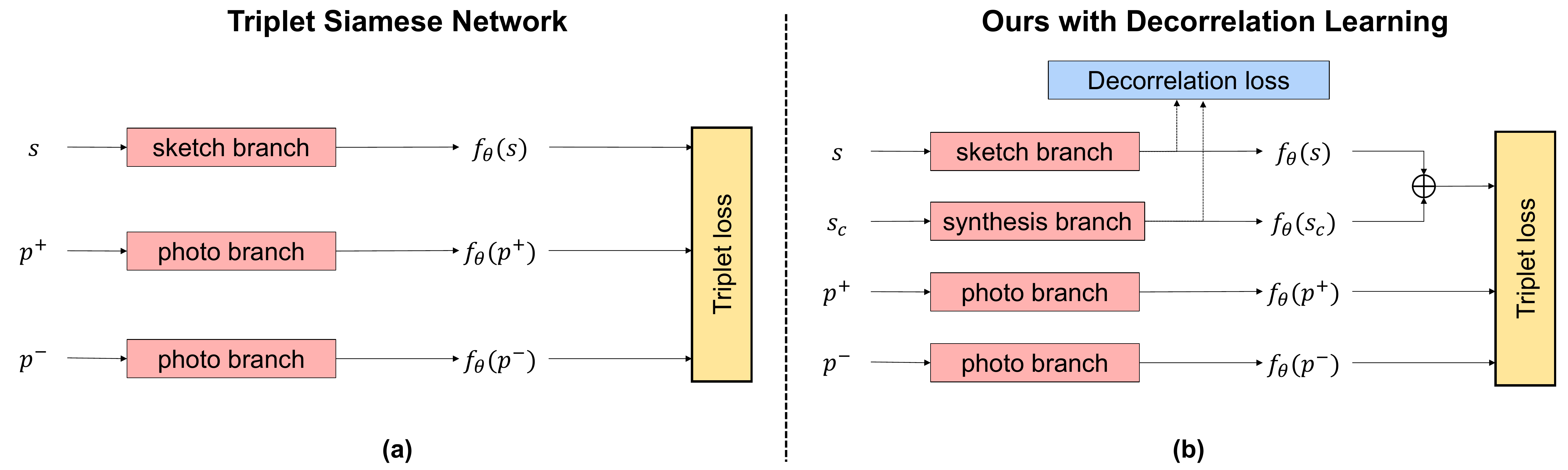}
\caption{(a) Existing three-branch Siamese Network\cite{yu2016sketch,sangkloy2016sketchy} vs. (b) our four-branch network with decorrelation loss. }
\label{fig:triplet}
\end{figure}

\keypoint{Problem definition:} For a given query sketch $s$ and a set of $N$ candidate photos $\{p_{i}\}_{i=1}^{N}\in P$, FG-SBIR aims to  find a specific photo containing the same instance as the query sketch. This can be solved by learning a joint sketch-photo embedding using a CNN $f_{\theta}$ \cite{yu2016sketch,sangkloy2016sketchy}. In this space, the visual similarity between a sketch $s$ and a photo $p$ can be measured simply as $D(s,p)=||f_{\theta}(s)-f_{\theta}(p)||_{2}^{2}$. 

\keypoint{Enforcing factorisation via de-correlation loss:}\quad  In our approach, clean and accurate contour features are already provided in $s_c$ via our style transfer network defined previously. Now we aim to extract detail-related features from $s$. To this end we introduce a decorrelation loss between $f_{\theta}(s)$ and $f_{\theta}(s_c)$:
\begin{equation}
\label{equ:ortho_loss}
\begin{aligned}
L_{decorr}=||\overbar{f_{\theta}(s)}^{T}\overbar{f_{\theta}(s_c)}||_{F}^2,
\end{aligned}
\end{equation}
\noindent where $\overbar{f_{\theta}(s)}$ and $\overbar{f_{\theta}(s_c)}$ are obtained by  normalising $f_{\theta}(s) $ and $f_{\theta}(s_c)$ with zero-mean and unit-variance respectively, and $||.||_{F}^2$ is the squared Frobenius norm. This ensures that $f_\theta(s)$ encodes detail-related features in order to meet the decorrelation constraint with complementary contour encoding $f_\theta(s_c)$.

\keypoint{Model design:} Existing deep FG-SBIR models \cite{yu2016sketch,pang2017fgsbir} adopt a three-branch Siamese network architecture, shown in Fig.~\ref{fig:triplet}(a). Given an anchor sketch $s$ and a positive photo $p^+$ containing the same object instance and a negative photo $p^-$, the outputs of the three branches are subject to a triplet ranking loss to align the sketch and photo in the discriminative joint embedding space learned by $f_{\theta}$. To exploit our contour and detail representation, we use a four-branch Siamese network with inputs $s, s_c, p^+, p^-$ respectively (Fig.~\ref{fig:triplet}(b)). The extracted features from $s$ and $s_c$ are then fused before being compared with those extracted from $p^+$ and $p^-$.  The fusion is denoted as $f_{\theta}(s)\oplus f_{\theta}(s_c)$, where $\oplus$ is the element-wise addition\footnote{Other fusion strategies have been tried and found to be inferior. }. The triplet ranking loss is then formulated as:
\begin{equation}
\label{equ:triplet_loss}
\begin{aligned}
L_{tri} = \max(0, \Delta + D(f_{\theta}(s)\oplus f_{\theta}(s_c),f_{\theta}(p^{+}))- D(f_{\theta}(s)\oplus f_{\theta}(s_c),f_{\theta}(p^{-})))
\end{aligned}
\end{equation}
\noindent where $\Delta$ is a hyperparameter representing the margin between the query-to-positive  and query-to-negative distances. Our final objective for discriminatively training SBIR becomes:
\begin{equation}
\label{equ:fgsbir_loss}
\begin{aligned}
\min_{\theta}\sum \mathop{}_{t\in T} L_{tri}+\lambda_{decorr}L_{decorr}
\end{aligned}
\end{equation}
\noindent we set $\Delta = 0.1, \lambda_{decorr}=1$ in our experiments so two losses have equal weights.

\section{Experiments}

\subsection{Experimental Settings}
\label{sec:exp setting}

\keypoint{Dataset and preprocessing:} We use the public QMUL-Shoe-V2 \cite{sketchx} dataset, the largest single-category paired sketch-photo dataset to date, to train and evaluate both our sketch style transfer model and FG-SBIR model. It contains $6648$ sketches and $2000$ photos. We follow its standard train/test split with $5982$ and $1800$ sketch-photo pairs respectively. Each shoe photo is annotated with $37$ part-based semantic attributes. We remove four decoration-related ones (`frontal', `lateral', `others' and `no decoration'), which are contour-irrelevant and keep the rest. Since our style transfer model is unsupervised and does not require paired training examples, we use a large shoe photo dataset UT-Zap50K dataset \cite{yu2014fine} as the target photo domain. This consists of  50,025 shoe photos which are disjoint with the QMUL-Shoe-V2 dataset. For training the style transfer model, we scale and centre the sketches and photo contours to $64 \times 64$ size, while for FG-SBIR model, the inputs of all four branches are resized to  $256 \times 256$. 

\keypoint{Photo contour extraction:} We obtain the contour $c$ from a photo $p$ as follows: (i) extracting edge probability map $e$ using \cite{ZitnickECCV14edgeBoxes} followed by non-max suppression; (ii) $e$ is binarised by keeping the edge pixels with edge probabilities smaller than $x$, where $x$ is dynamically determined so that when $e$ contains many non-zero edge pixel detections, $x$ should be small to eliminate the noisy ones, e.g., texture.  This is achieved by formulating $x = e_{sort}(l_{sort} \times \min(\alpha e^{-\beta \times r}, 0.9))$, where $e_{sort}$ is the edge pixels detected in $e$ sorted in the ascending order, $l_{sort}$ is the length of $e_{sort}$, and $r$ is the ratio between detected and total pixels. We set $\alpha=0.08, \beta=0.12$ in our experiments. Examples of photos and their extracted contours can be seen in the last two columns of Fig.~\ref{fig:comp}.

\keypoint{Implementation details:} We implement both models in Tensorflow with a single NVIDIA $1080$Ti GPU. For the \textbf{style transfer task}: as illustrated in Fig.~\ref{fig:framework}, we denote $k\ast k$ conv as a $k \times k$ Convolution-BatchNorm-ReLU layer with stride 1 and $k\ast k$ residual as a residual block that contains two $k\ast k$ conv blocks with reflection padding to reduce artifacts. Upscale operation is performed with bilinear up-sampling. We do not use BatchNorm and replace ReLU activation with Tanh for the last output layer. Our discriminator has the same architecture as in \cite{isola2016image}, but with BatchNorm replaced with LayerNorm \cite{ba2016layer} since gradient penalty is introduced. The number of discriminator iterations per generator update is set as $1$. We trained for $50k$ iterations with a batch size of $64$. For the \textbf{FG-SBIR task}: we fine-tune ImageNet-pretrained ResNet-50 \cite{he2016deep} to obtain $f_{\theta}$  with the final classification layer removed. Same with \cite{yu2016sketch}, we enforce $l_2$ normalisation on $f_{\theta}$ for stable triplet learning. We train for $60k$ iterations with a triplet batch size of $16$. For both tasks, the Adam \cite{kingma2014adam} optimiser is used, where we set $\beta_1=0.5$ and $\beta_2=0.9$ with an initial learning rate of $0.0001$ respectively.

\keypoint{Competitors:}  For style transfer, four competitors are compared. \textbf{Pix2pix} \cite{isola2016image} is a supervised image-to-image translation model. It assumes that visual connections can be directly established between sketch and contour pairs with $l_1$ translation loss and adversarial training. Note that we can only use the QMUL-Shoe-V2 train split for training Pix2pix, rather than UT-Zap50K, since sketch-photo pairs are required. \textbf{UNIT} \cite{liu2017unsupervised} is the latest variant of the popular unsupervised  CycleGAN \cite{zhu2017unpaired, kim2017learning, yi2017dualgan}. Similar to our model, it also has a shared embedding construction subnet. Unlike our model, there is no attribute prediction regularisation and visual consistency instead of embedding consistency is enforced.  \textbf{UNIT-vgg}: for fair comparison, we substitute the learned-from-scratch encoder in UNIT to our fixed VGG-encoder, and introduce the same self-residual architecture in the decoder.  \textbf{Ours-attr}: This is a variant of our model without the attribute prediction task for embedding regularisation. For FG-SBIR, competitors include: \textcolor{black}{\textbf{Sketchy} \cite{sangkloy2016sketchy} is a three-branch Heterogeneous triplet network. For fair comparison, the same ResNet50 is used as the base network. \textbf{Vanilla-triplet} \cite{yu2016sketch} differs from Sketchy in that a Siamese architecture is adopted. It is vanilla as the model is trained without any synthetic augmentation.} \textbf{DA-triplet}\cite{song2018sketch} is the state-of-the-art model, which uses  synthetic sketches from photos as a means of data augmentation to pretrain the Vanilla-triplet network and fine-tune it with real human sketches. \textbf{Ours-decorr} is a variant of our model, obtained by discarding the decorrelation loss.

\subsection{Results on Style Transfer}

\begin{figure}[t]
\centering
\includegraphics[width=0.85\textwidth]{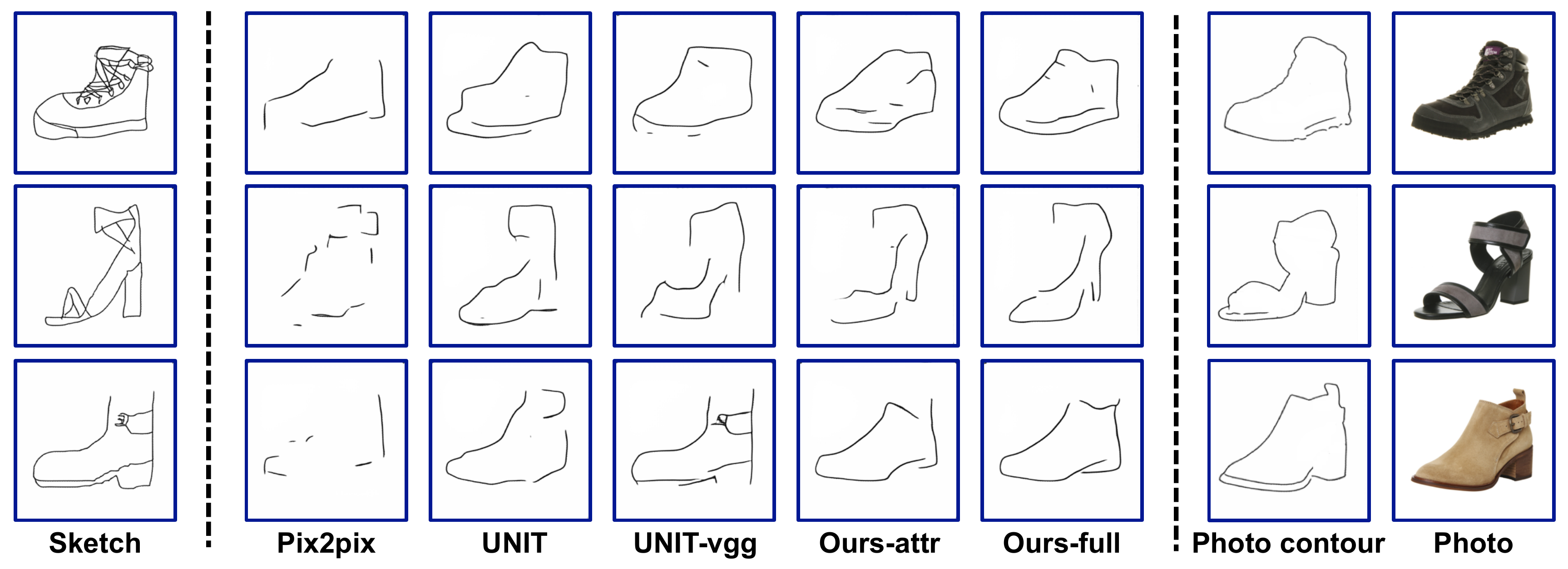}
\caption{Different competitors for translating sketching abstraction at contour-level. Illustrations shown here have never been seen by its corresponding model during training. }
\label{fig:comp}
\end{figure}

\begin{figure}[t]
\centering
\includegraphics[width=0.9\textwidth]{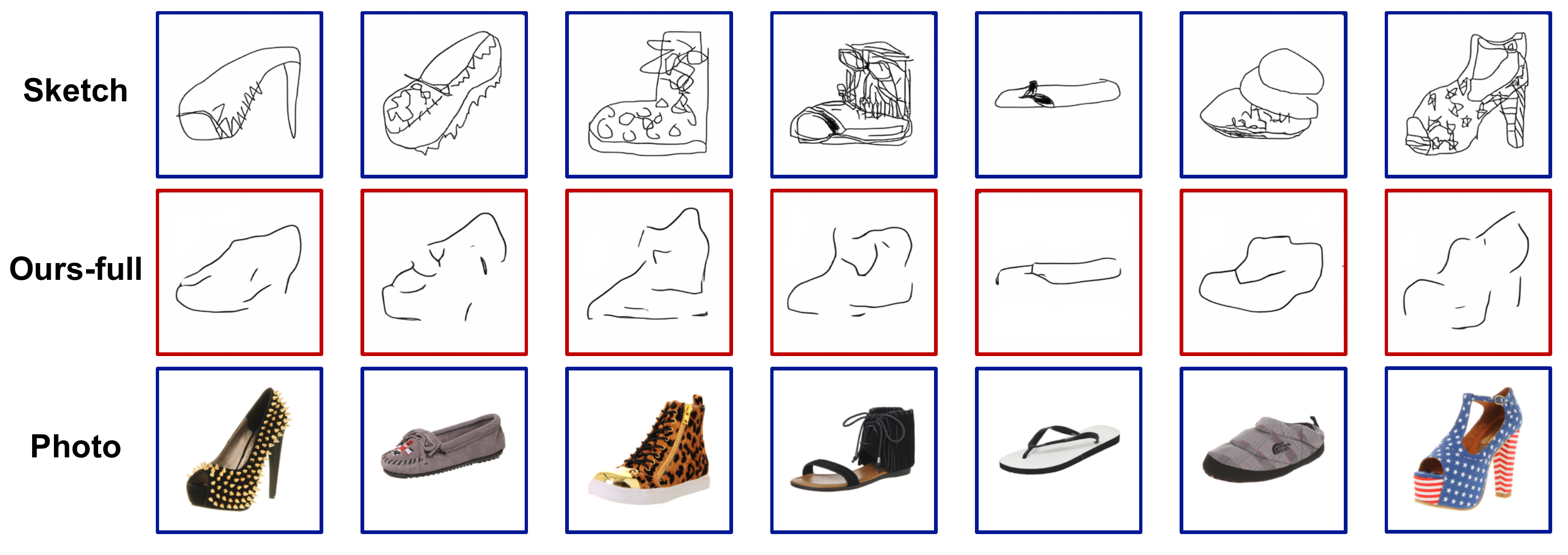}
\caption{Typical failure of our model when sketching style is too abstract or complex.}
\label{fig:failure}
\end{figure}

\keypoint{Qualitative results:} Fig.~\ref{fig:comp} shows example synthesised sketches using the various models.  It shows clearly that our method is able to invert the sketching process by effectively factorising out any details inside the object contour and restyling the remaining contour parts with smooth strokes and more realistic perspective geometry. In contrast, the supervised model Pix2pix failed completely due to sparse training data and the assumption of pixel-to-pixel alignment across the two domains. The unsupervised  UNIT model is able to remove the details, but struggles to emulate the style of the object photo contours featured with smooth and continuous strokes. Using a fixed VGG-16 as encoder (UNIT-vgg) alleviates the problem but introduces the new problem of keeping the detail part. These results suggest that the visual cycle  consistency constraint used in UNIT is too strong a constraint on the embedding subnet, leaving it with little freedom to perform both the detail removal and contour restyling tasks. As an ablation, we compare ours-attr with ours-full and observe that the attribute prediction task does provide a useful regularisation to the embedding subnet to make the synthesised contour more smooth and less fragmented. Our model is far from being perfect. Fig.~\ref{fig:failure} shows some failure cases. Most failure cases are caused by the sketcher unsuccessfully attempting to depict objects with rich texture by an overcomplicated sketch. This suggests that our model is mostly focused on the shape cues contained in sketches and confused by the sudden presence of large amounts of texture cues. 

\keypoint{Quantitative results:} 
Quantitative evaluation of image synthesis models remains an open problem. Consequently, most studies either run human perceptual studies or explore computational metrics attempting to predict human perceptual similarity judgements \cite{salimans2016improved, heusel2017gans}. We perform both quantitative evaluations. 
\noindent\textbf{Computational evaluation:} In this evaluation, we seek a metric based on the insight that if the synthesised sketches are realistic and free of distortion, they should be useful for retrieving photos containing the same objects, despite the fact that the details inside the contours may have been removed.  We thus retrain the FG-SBIR model of \cite{yu2016sketch} on the QMUL-Shoe-V2 training split and used the  sketches synthesised using different style transfer models  to retrieve photos in QMUL-Shoe-V2 test split. The results in Table \ref{tab:syn_retrieval} show that  our full model outperforms all competitors. The performance gap over the chance suggests that despite lack of detail, our synthetic sketches still capture instance-discriminative visual cues. The superior results to the competitors indicate the usefulness of cyclic embedding consistency and attribute prediction regularisation.
\noindent\textbf{Human perceptual study:}\quad We further evaluate our model via a human subjective study. We recruit $N$ ($N=10$) workers and ask each of them to perform the same pairwise $A/B$ test based on the $50$ randomly-selected sketches from QMUL-Shoe-V2 test split. Specifically, each worker undertakes two trials, where three images are given at once, i.e., a sketch and two restyled version of the sketch using two compared models.  The worker is then asked to choose one synthesised sketch based on two criteria: (i) correspondence (measured as $r_c$): which image keeps more key visual traits of the original sketches, i.e., more instance-level identifiable; (ii) naturalness (measured as $r_n$): which image looks more like a contour extracted from a  shoe photo. The left-right order and the image order are randomised to ensure unbiased comparisons. We denote each of the $2N$ ratings for each synthetic sketch under one comparative test as $c_i$ and $n_i$ respectively, and compute the correspondence measure $r_c =\sum_{i=1}^{N} c_i$, and naturalness measure $r_n=\sum_{i=1}^{N} n_i$. We then  average them to obtain one score based on a weighting: $r_{avr}= \frac{1}{N}(w_c r_c + w_n r_n)$. Intuitively, $w_c$ should be greater than $w_n$ because ultimately we care more about how the synthesised sketches help FG-SBIR. In Table \ref{tab:human_study}, we list in each cell the percentage of trials where our full model is preferred over the other competitors. Under different weighting combinations, the superiority of our design is consistent ($>50$\%), drawing the same conclusion as our computational evaluation. In particular, compared with prior state-of-the-art, UNIT, our full model is preferred by humans nearly $90\%$ of the time.

\begin{table}[tb]
\centering
\resizebox{0.8\columnwidth}{!}{
{\renewcommand{\arraystretch}{1.2}
\begin{tabular}{l<{\hspace{0.35pc}}|c<{\hspace{0.35pc}}|c<{\hspace{0.35pc}}|c<{\hspace{0.35pc}}|c<{\hspace{0.35pc}}|c<{\hspace{0.35pc}}|c}
\hline
 &Chance& Pix2pix \cite{yu2016sketch} & UNIT \cite{liu2017unsupervised} & UNIT-vgg & Ours-attr & Ours-full \\
\hline
acc@1 & 0.50\% & 3.60\%  & 4.50\%  & 4.95\% & 6.46\% & {\bf 8.26\%}  \\
\hline
acc@5 & 2.50\% & 10.51\%  & 15.02\%  & 17.87\% & 22.22\% & {\bf 23.27\%}  \\
\hline
acc@10 & 5.00\% & 17.87\% & 26.28\%  & 29.88\% & 31.38\% & {\bf 35.14\%} \\
\hline
\end{tabular}}
}
\protect\caption{Comparative retrieval results  using the synthetic sketches obtained using different models.}
\centering
\label{tab:syn_retrieval}

\resizebox{0.8\columnwidth}{!}{
{\renewcommand{\arraystretch}{1.2}
\begin{tabular}{c<{\hspace{0.3pc}}|c|c|c}
\hline
$(w_c, w_n)$ & UNIT vs. Ours-full & UNIT-vgg vs. Ours-full & Ours-attr vs. Ours-full \\
\hline
(0.9, 0.1) & 88.0\% &72.0\% &62.0\% \\
\hline
(0.8, 0.2) & 88.0\% &70.0\% &64.0\% \\
\hline
(0.7, 0.3) & 88.0\% &70.0\% &64.0\% \\
\hline
(0.6, 0.4) & 86.0\% &68.0\% &62.0\% \\
\hline
(0.5, 0.5) & 84.0\% &70.0\% &64.0\% \\
\hline
\end{tabular}}
}
\protect\caption{Pairwise comparison results of human perceptual study. Each cell lists the percentage where our full model is preferred over the other method. Chance is at $50\%$.}
\centering
\label{tab:human_study}

\resizebox{0.8\columnwidth}{!}{
{\renewcommand{\arraystretch}{1.2}
\begin{tabular}{c<{\hspace{0.3pc}}|c<{\hspace{0.3pc}}|c<{\hspace{0.3pc}}|c<{\hspace{0.3pc}}|c<{\hspace{0.3pc}}}
\hline
Sketchy \cite{sangkloy2016sketchy} &Vanilla-triplet \cite{yu2016sketch} &DA-triplet \cite{song2018sketch} & Ours-decorr  & Ours-full \\
\hline
 21.62\% &33.48\% & 33.78\%  & 33.93\%  & {\bf 35.89\%}  \\
\hline
\end{tabular}}
}
\protect\caption{Comparative results on QMUL-Shoe-V2. Retrieval accuracy at rank 1 (acc@1).}
\centering
\label{tab:fgsbir}
\end{table}

\subsection{Results on FG-SBIR}

\keypoint{Quantitative:}
In Table \ref{tab:fgsbir}, we compare the proposed  FG-SBIR model (Ours-full) with three state-of-the-art alternatives (Sketchy, Vanilla-triplet and DA-triplet) and a variant of our model (Ours-decorr). The following observations can be made: (i) Compared with the three existing models, our full model yields 14.27\%, 2.41\% and 2.11\% acc@1 improvements respectively. Given that the three competitors have exactly the same base network in each network branch, and the same model complexity as our model, this demonstrates the effectiveness of our complementary detail representation from contour-detail factorisation. (ii) Without the decorrelation loss, Ours-decorr produces similar accuracy as the two baselines and is clearly inferior to Ours-full. This is not surprising -- without forcing the original sketch ($s$) branch to extract something different from the sketch contour ($s_c$) branch (i.e., details), the fused features will be dominated by the $s$ branch as $s$ contains much richer information. The four-branch model thus degenerates to a three-branch model.   

\begin{figure}[t]
\centering
\includegraphics[width=0.85\textwidth]{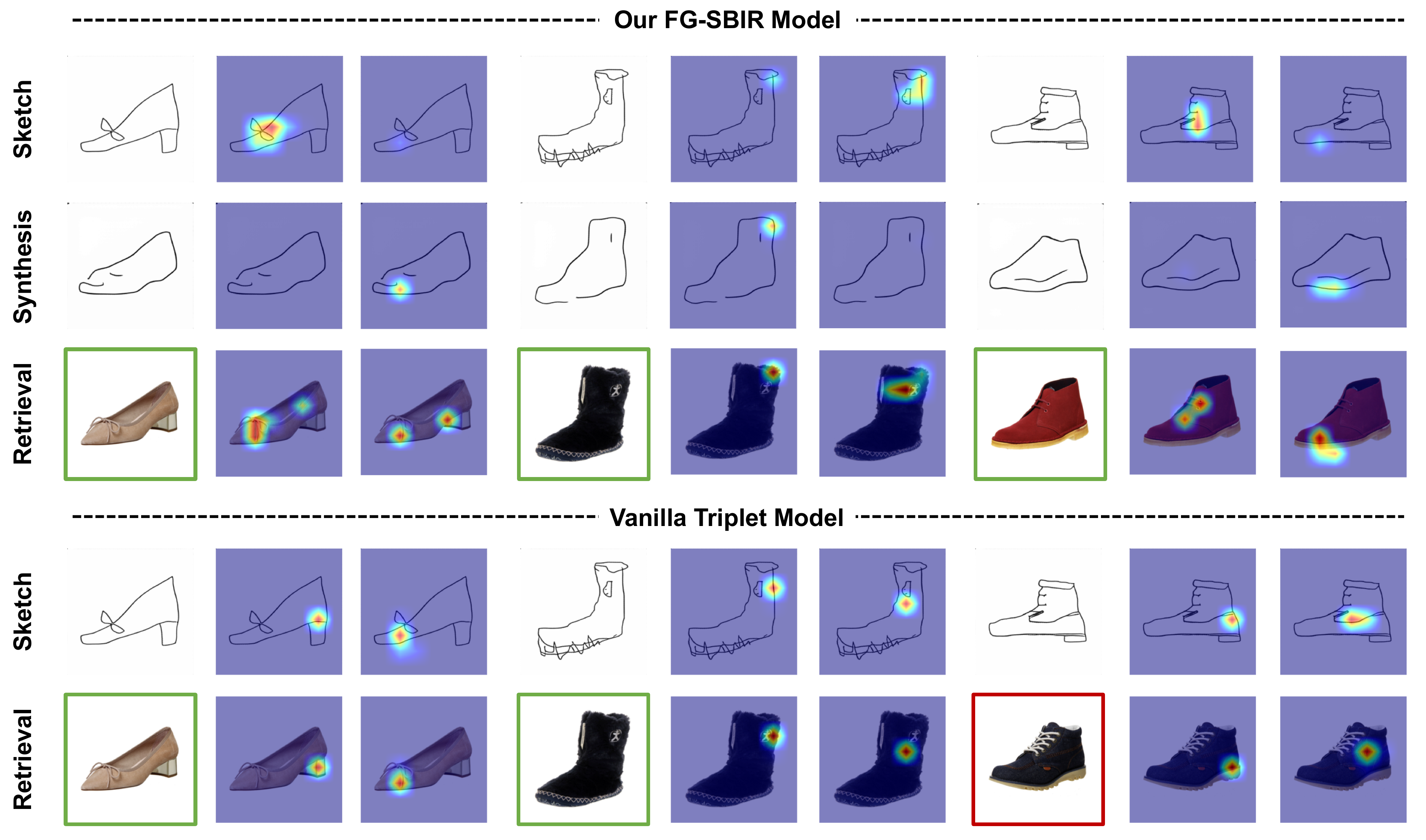}
\caption{We highlight supporting regions for the top 2 most discriminative feature dimensions of two compared models. Green and red borders on the photos indicate correct and incorrect retrieval, respectively.}
\label{fig:vis}
\end{figure}

\keypoint{Visualisation:}
We carry out model visualisation to demonstrate that $f_{\theta}(s)$ and $f_{\theta}(s_c)$ indeed capture different and complementary features that are useful for FG-SBIR, and give some insights on why such a factorisation helps. To this end,  we use Grad-Cam \cite{selvaraju2016grad} to highlight where in the image the discriminative features are extracted using our model. Specifically, the two non-zero dimensions of $f_{\theta}(s)\oplus f_{\theta}(s_c)$ that contribute the most similarity for the retrieval are selected and their gradients are propagated back along the $s$ and $s_c$ branches as well as the photo branch to locate the support regions. The top half of Fig.~\ref{fig:vis} shows clearly that (i) the top discriminative features are often a mixture of contour and detail as suggested by the highlighted regions on the photo images; and (ii) the corresponding regions are accurately located in $s$ and $s_c$; importantly the contour features activate mostly in $s_c$ and detail features in $s$. This validates that factorisation indeed takes place. In contrast, the bottom half of Fig.~\ref{fig:vis} shows that using the vanilla-triplet model without the factorisation, the model appears to be overly focused on the details, ignoring the fact that the contour part also contains useful information for matching object instances. This leads to failure cases (red box) and explains the inferior performance of vanilla-triplet.

\section{Conclusion}

We have for the first time proposed a framework for inverting the iconic rendering process in human free-hand sketch, and for contour-detail factorisation. Given a sketch, our deep style transfer model learns to factorise out the details inside the object contour and invert the remaining contours to match more geometrically realistic contours extracted from photos. We subsequently develop a sketch-photo joint embedding  which completes the inversion process by extracting distinct complementary detail features for FG-SBIR. We demonstrated empirically that our style transfer model is more effective compared to existing models thanks to a novel cyclic embedding consistency constraint. We also achieve state-of-the-art FG-SBIR results by exploiting our sketch inversion and factorisation.

%
%
%
 \bibliographystyle{splncs04}
 \bibliography{egbib}
%
%
%
%
\end{document}